\documentclass[conference]{IEEEtran}
\IEEEoverridecommandlockouts
\usepackage{cite}
\usepackage{amsmath,amssymb,amsfonts}
\usepackage{algorithmic}
\usepackage{graphicx}
\usepackage{textcomp}
\usepackage{xcolor}
\usepackage{booktabs}
\usepackage{multirow} 
\usepackage{array}
\usepackage{tabularx}
\usepackage{booktabs}
\def\BibTeX{{\rm B\kern-.05em{\sc i\kern-.025em b}\kern-.08em
    T\kern-.1667em\lower.7ex\hbox{E}\kern-.125emX}}
\begin{document}

\title{Group-CLIP Uncertainty Modeling for Group Re-Identification
\thanks{$^{1}$ Equal Contribution. $^{*}$ Corresponding authors.}
}

\author{\IEEEauthorblockN{Qingxin Zhang$^1$}
\IEEEauthorblockA{\textit{Glasgow College Hainan, UESTC} \\
\textit{University of Electronic Science and Technology} \\
\textit{of china} \\
Hainan, China\\
2022360902033@std.uestc.edu.cn}
\and
\IEEEauthorblockN{Haoyan Wei$^1$}
\IEEEauthorblockA{\textit{School of Electronic Information} \\
\textit{Sichuan University}\\
Chengdu, China \\
hywei@stu.scu.edu.cn}
\and
\IEEEauthorblockN{Yang Qian*}
\IEEEauthorblockA{\textit{USC Viterbi School of Engineering} \\
\textit{University of Southern California}\\
Los Angeles, California, US \\
yqian442@usc.edu.cn}
% \and
% \and
% \and
}

\maketitle

\begin{abstract}
Group Re-Identification (Group ReID) aims matching groups of pedestrians across non-overlapping cameras. Unlike single person ReID, Group ReID focuses more on the changes in group structure, emphasizing the number of members and their spatial arrangement. However, most methods rely on certainty-based models, which consider only the specific group structures in the group images, often failing to match unseen group configurations. To this end, we propose a novel Group-CLIP Uncertainty Modeling (GCUM) approach that adapts group text descriptions to undetermined accommodate member and layout variations. Specifically, we design a Member Variant Simulation (MVS) module that simulates member exclusions using a Bernoulli distribution and a Group Layout Adaptation (GLA) module that generates uncertain group text descriptions with identity-specific tokens. In addition, we design a Group Relationship Construction Encoder (GRCE) that uses group features to refine individual features, and employ cross-modal contrastive loss to obtain generalizable knowledge from group text descriptions. It is worth noting that we are the first to employ CLIP to Group ReID, and extensive experiments show that GCUM significantly outperforms state-of-the-art Group ReID methods.
\end{abstract}


\section{Introduction}
Group Re-Identification \cite{r9, r25, r40} aims to identify and retrieve groups with the same identity from gallery images based on specific queries. There are few studies on Group ReID at present, because it not only faces traditional challenges\cite{r41, r42} like occlusion, clothing changes and lighting variations but also additional issues such as group members (i.e., the number of group members may change across different cameras) and layout (i.e., the spatial positions of group members may differ across various cameras) changes. Existing Group ReID methods \cite{r8,r10,r11,r12,r24} mainly model finite group structures from the dataset using non-augmentation strategies, which leads to poor performance in the face of uncertain group layouts with real-world variations.

\begin{figure}[!t]
\centering
\includegraphics[width=1\linewidth]{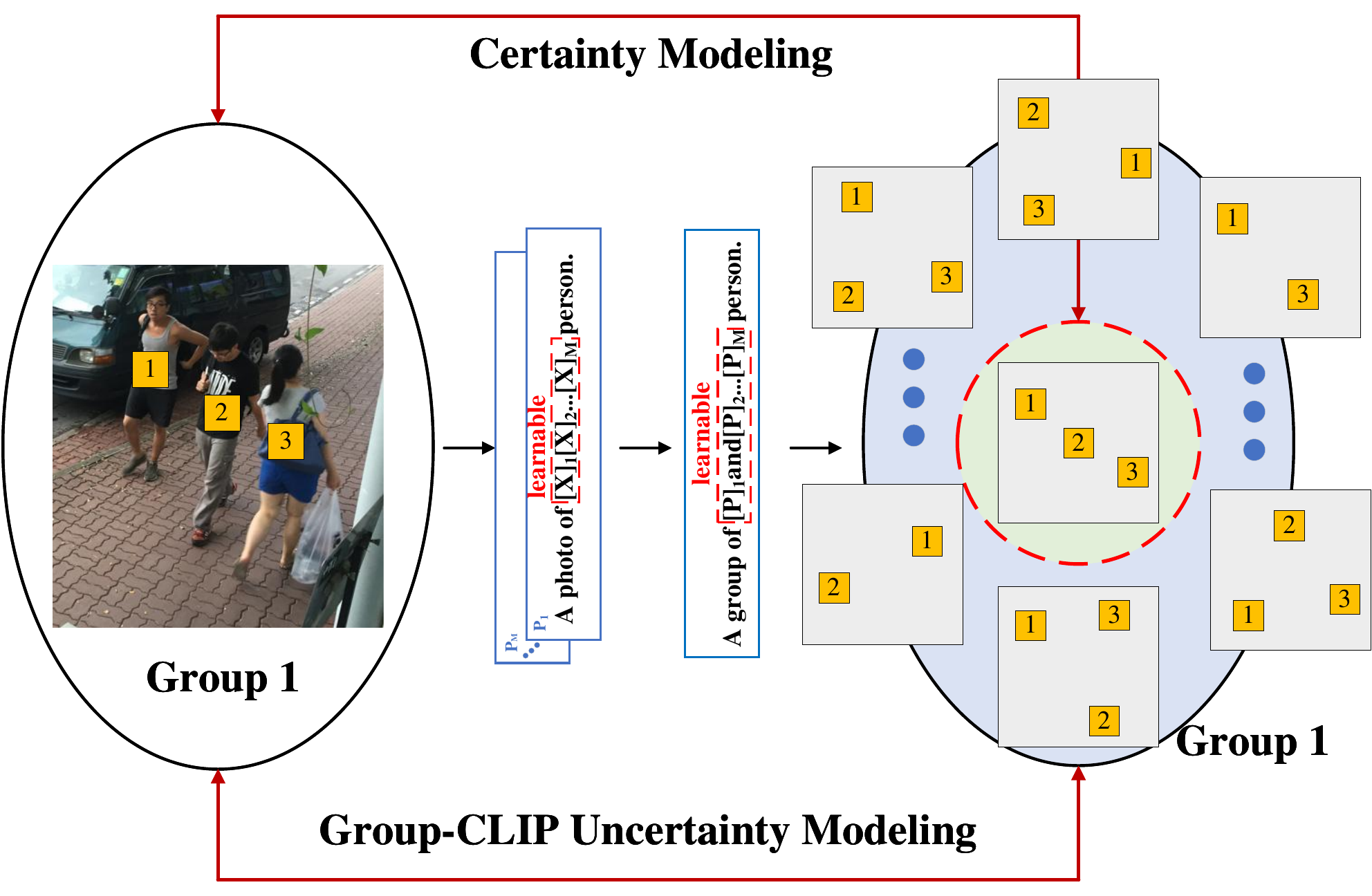}
\caption{The motivation of our GCUM. (1) Certainty modeling learns from finite group structures; (2) Text descriptions can effectively adapt to variations in group members and layouts, enabling the learning of a broader range of group structures.}
\label{fig1}
\end{figure}

\begin{figure*}[t]
\centering
\includegraphics[width=1\linewidth]{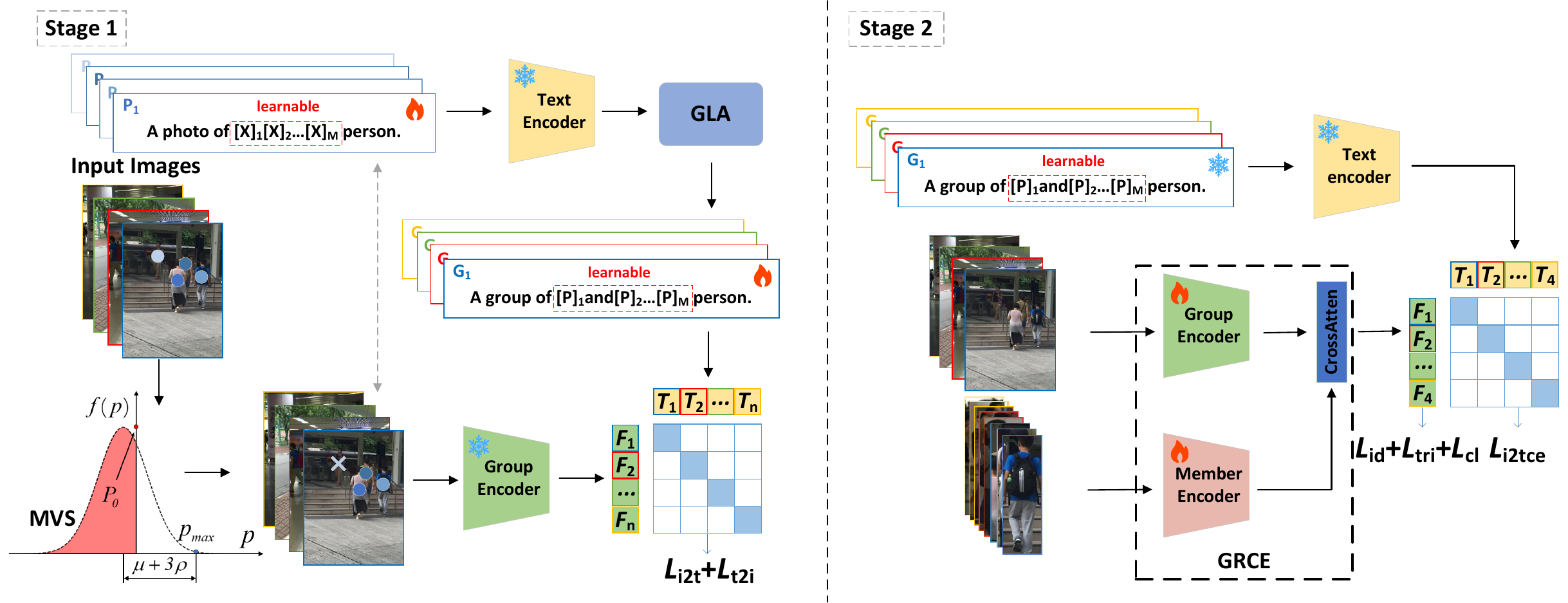}
\caption{A schematic diagram of the GCUM training process. (1) Stage 1: Utilizing the Member Variant Simulation and Group Layout Adaptation module to generate robust group text descriptions that can adapt to changes in group members and layout; (2) Stage 2: Using the group text descriptions generated in the first stage to guide the learning of broader group structures in the visual features.}
\label{fig2}
\end{figure*}

Recently, CLIP-ReID \cite{r21} introduced the use of Visual-Language Pretraining (VLP) \cite{r36, r37, r38, r39} for person Re-Identification (ReID) by generating general text descriptions for each identity. This approach utilizes multi-modal descriptions to enhance the model's capability to distinguish between pedestrians. However, CLIP-ReID struggles with group identification, as it fails to capture the structure and potential associations among group members. Thus, accurately modeling complex group structures and implicit relationships within groups using CLIP presents a significant challenge. To address this, we introduce uncertain implicit expressions into group text descriptions, such as "\textit{A group of people in red shirts, blue trousers, black shoes, and possibly blue shirts, black trousers, brown shoes.}" These descriptions offer clues for recognizing individual appearances (e.g., clothing and accessories) without the need to adjust for changes in layout. Moreover, they can adapt to variations in group size by incorporating phrases like "\textit{potentially present}" to account for different group compositions. As illustrated in Fig. \ref{fig1}, these uncertain text descriptions, which account for member and layout variability, enable the identification of a broader range of group structures.

To address the challenges in group ReID, we propose a novel Group-CLIP Uncertainty Modeling (GCUM) approach, which utilizes uncertain group text descriptions to handle variations in group members and layouts. Our method employs a two-stage training process to enhance the recognition of unseen group structures. In the first stage, we introduce a Group Layout Adaptation (GLA) module to generate uncertain group text descriptions using identity-specific tokens. This module incorporates terms like “\textit{potentially present}” to account for variations in group composition. Additionally, we implement a Member Variant Simulation (MVS) module that employs a Bernoulli distribution to randomly exclude certain group members, simulating diverse group scenarios. In the second stage, we develop a Group Relationship Construction Encoder (GRCE) to refine and aggregate individual features. This encoder filters out irrelevant members and enhances group identification by leveraging cross-modal contrastive loss, which transfers knowledge from text features and adapts to changes in group members and layouts.

The following are the primary contributions of this paper:
\begin{itemize}
    \item To our knowledge, we are the first to apply CLIP to Group ReID and propose a novel GCUM approach to effectively adapt to diverse group structures.
    \item We design a Group Layout Construction module and a Member Variation Simulation module to generate ambiguous group text descriptions and add “potentially present” terms to the descriptions, respectively.
    \item Group Relationship Construction Encoder is proposed to fully utilize the generalizable knowledge from group text descriptions regarding member and layout variations.
\end{itemize}

\section{Method}
As shown in Fig. \ref{fig2}, our proposed GCUM approach comprises MVS, GLA and GRCE modules. Detailed descriptions are provided in the following subsections.
\subsection{Member Variant Simulation Module}
For each group image ${x}_{i}$, a binary mask ${{\overrightarrow{m}}_{i}}$ is constructed to either retain or remove individual pedestrians from the group image. The specific construction of the mask ${{\overrightarrow{m}}_{i}}$ is as follows:
\begin{equation}
\begin{aligned}
{{\overrightarrow{m}}_{i}}(j)\sim \beta (\text{1}-p),\text{ }s.t.\text{ }p\sim N(\mu ,\sigma ;{{p}_{\text{0}}},{{p}_{max}}),\
\end{aligned}
\end{equation}
where $\beta(\cdot)$ represents a Bernoulli distribution, where the value is 0 with a probability of $p$ and 1 with a probability of $\text{1}-p$. The value ${{\overrightarrow{m}}_{i}(j)}$ is the mask value corresponding to the $j$-th individual in the $i$-th group image, where 0 and 1 indicate the removal or retention of the individual, respectively. ${p}_{\text{0}}$ and ${{p}_{max}}$ are prior attributes related to group properties. The number of remaining members involved in the group class token learning can be represented as $\left| \overrightarrow{m_i} \right|={{\sum\nolimits_{j}{\overrightarrow{m}}}_{i}(j)}$.

In addition, we provide a learnable quantity relationship matrix ${{E}^{m}}\in {{\mathbb{R}}^{{M}_{0}\times {dim}}}$ to describe the relationships among different group members. The submatrix formed by the first $k$ columns encodes the numerical representation of $k$ members, where ${M}_{0}$ is the preset maximum number of group members, and $dim$ is the feature embedding dimension. The steps to simulate variations in group size are as follows:
\begin{equation}
    \begin{split}
        & {{S}_{i}}^{\prime }={{S}_{i}}\odot {{\overrightarrow{m}}_{i}}, \\ 
        & {{\overrightarrow{t}}_{s}}^{\prime }={{\overrightarrow{t}}_{s}}+{{E}_{m}}(\text{top}(|\overrightarrow{m}|))*{{S}_{i}}^{\prime }, \\ 
        & {{F}_{i}}=Concat({{\overrightarrow{t}}_{s}}^{\prime },{{S}_{i}}^{\prime }), \\
    \end{split}
\end{equation}
where $S_{\text{i}}$ is the sequence of features from the current group members, and $\odot$ denotes the filtering operation. a represents selecting the first  elements. $\text{top}(k)$ indicates selecting the first $k$ elements, $\overrightarrow{t_{s}}$ denotes the group class token, and $*$ represents the matrix multiplication operation. Finally, the refined group class token $\overrightarrow{t}_{s}^{\prime }$ and the filtered member sequence 
${{S}_{i}}^{\prime }$ are concatenated and used as input to the remaining transformer layers.

\subsection{Group Layout Adaptation Module}
Group text descriptions can effectively adapt to various member layouts. However, generating text in a coarse manner for each group may overlook some member-specific cues and be impacted by variations in the number of members within the same group. Therefore, we leverage CLIP's powerful cross-modal semantic construction capabilities to generate preliminary text descriptions for each member within the group. Then, we design the Group Layout Adaptation (GLA) module to aggregates member text descriptions to form a comprehensive group structure description.

To improve the robustness of group text descriptions against variations in member count, we use the MVS module to add “potentially present” conditional options. Specifically, the cross-modal contrastive loss connects visual features (with varying member counts) to text features (with a fixed count set to a predefined maximum), adaptively incorporating conditional options into the descriptions. The calculation of the cross-modal contrastive loss is as follows:
\begin{equation}
\begin{aligned}
{{\mathcal{L}}_{t2i}}({{y}_{i}})=\frac{-\text{1}}{|p({{y}_{i}})|}\sum\limits_{p\in P({{y}_{i}})}{log\frac{exp\text{(}s\text{(}{{V}_{p}},{{T}_{{{y}_{i}}}}\text{))}}{\sum\nolimits_{\alpha =\text{1}}^{B}{exp(s({{V}_{a}},{{T}_{{{y}_{i}}}}))}}},
\end{aligned}
\end{equation}
%其中,a表示当前batch中与T属于正样本的索引集合，B是batch中样本的总数，|a|是a的基数。
where $P({{y}_{i}})=\{p\in \text{1}\ldots B:{{y}_{p}}={{y}_{i}}\}$ represents the index set of positive samples in the current batch that belong to $T_{y_{i}}$, $B$ is the total number of samples in the batch, and $|\cdot|$ is the cardinality value. The calculation process of $\mathcal{L}_{t2i}$ is similar to this one.

\subsection{Group-CLIP}

%我们为组内的每个个体都引入了M个可学习token来学习身份辨别描述，这些token接着被嵌入到“A photo of a [X]1[X]2[X]3...[X]M person”句式中，其中每一个a都是具有相同维度的可学习token。接着，借助我们设计的GLA将属于同一组的tokens聚合成组文本描述，表示为“A group of ${{[P]}_{\text{1}}}{{[P]}_{\text{2}}}{{[P]}_{3}}\ldots {{[P]}_{\text{K}}}$ persons”.其中K为组内定义最大成员数，${{[p]}_{k}}(k\in \text{1},\ldots K)$为第k个组内成员的token集合。个体和组文本描述都输入到文本编码器中捕获文本特征，分别与个体和组视觉特征计算跨模态对比损失，用于更新可学习的token向量。值得注意的是为了增强组文本描述对成员数量改变的适应性，我们MVS模拟组视觉特征中的数量变化。第一阶段的训练损失可以表述为：
\textbf{The first training stage.} We introduce $M$ learnable tokens for each individual within the group to capture identity-discriminative descriptions. These tokens are then embedded into the sentence “A photo of a ${{[X]}_{\text{1}}}{{[X]}_{\text{2}}}{{[X]}_{3}}\ldots {{[X]}_{M}}$ person”, where each ${{[X]}_{m}}(m\in \text{1},\ldots M)$ is a learnable token with the same dimensionality. Then, using our designed GLA module, the tokens belonging to the same group are aggregated into a group text description, represented as “A group of ${{[P]}_{\text{1}}}{{[P]}_{\text{2}}}{{[P]}_{3}}\ldots {{[P]}_{K}}$ persons”, where $K$ denotes the predefined maximum number of members in the group, and ${{[P]}_{k}}(k\in \text{1},\ldots K)$ represents the token set for the $k$-th group member. Both individual and group text descriptions are input into the text encoder to capture text features. These features are used to compute cross-modal contrastive loss with individual and group visual features, updating the learnable token vectors. To improve the adaptability of group text descriptions to member count changes, our MVS module simulates variations in group member numbers. The training loss for the first stage can be expressed as:
\begin{equation}
{\mathcal{L}_{\text{stage1}}}={\mathcal{L}_{\text{i2t}}}+{\mathcal{L}_{\text{t2i}}}.
\end{equation}

\textbf{The second training stage.} During this stage, only our Group Relationship Construction Encoder (GRCE) is updated. Specifically, the group encoder extracts the group visual feature $V_{i}$, while the member encoder extracts the visual feature set ${S}_{i}^{\prime }$ for each individual in the group. Cross-attention is then used to refine and aggregate these individual features, producing an enhanced group visual feature ${{{V}'}_{i}}$. The refinement process is outlined as follows:
\begin{equation}
\begin{aligned}
{{{V}'}_{i}}=CrossAttention({{W}_{q}}({{V}_{i}}),{{W}_{k}}({{S}_{i}^{\prime }}),{{W}_{v}}({{S}_{i}^{\prime }})),
\end{aligned}
\end{equation}
where $W_{q}(\cdot)$, $W_{k}(\cdot)$ and $W_{v}(\cdot)$ are the matrices used to compute the query, key, and value, respectively. We optimize ${{V}'}_{i}$ using the triplet loss $\mathcal{L}_{tri}$ and identity loss $\mathcal{L}_{id}$ with label smoothing. The formulas for these losses are as follows:
\begin{equation}
\begin{split}
     & \mathcal{L}_{id}=\sum\limits_{k=\text{1}}^{N}-{{q}_{k}}log\text{(}{{p}_{k}}\text{)}, \\
    &\mathcal{L}_{tri}=max({{d}_{p}}-{{d}_{n}}+\alpha ,\text{0}), \\
\end{split}
\end{equation}
where $q$ represents the smoothed identity labels, $p$ denotes the predicted scores obtained by applying ${{V}'}_{i}$ through the classifier, and $k$ indicates the identity index. $d_{p}$ and $d_{n}$ represent the feature distances of positive and negative pairs, respectively, and $\alpha$ denotes the margin for the triplet loss $L_{tri}$.

For each group image, we calculate the image-to-text cross-entropy loss $L_{i2tce}$ using the group text descriptions obtained in the first training stage, as shown in Eq. (8). Note that after $L_{id}$, we apply label smoothing to $q_{k}$ in $L_{i2tce}$:
\begin{equation}
\begin{aligned}
{\mathcal{L}_{\text{i2tce}}}(i)=\sum\limits_{k=\text{1}}^{N}{-{{q}_{k}}log\frac{exp(s({{{V}'}_{i}},{{T}_{{{y}_{K}}}}))}{\sum\nolimits_{{{y}_{a}}=\text{1}}^{N}{exp(s({{{V}'}_{i}},{{T}_{{{y}_{a}}}}))}}}.
\end{aligned}
\end{equation}

Finally, the loss functions used in the second training stage are summarized as follows:
\begin{equation}
\begin{aligned}
{\mathcal{L}_{\text{stage}2}}={\mathcal{L}_{\text{id}}}+{\mathcal{L}_{\text{tri}}}+{\mathcal{L}_{\text{i2tce}}}.
\end{aligned}
\end{equation}

\begin{table*}[h!]
\centering
\caption{Comparison of the proposed method with the state-of-the-arts on RoadGroup, iLIDS-MCTS, and CSG}
\begin{tabular}{lcccccccccccc}
\toprule
\multirow{2}{*}{Method} & \multicolumn{4}{c}{RoadGroup} & \multicolumn{4}{c}{iLIDS-MCTS} & \multicolumn{4}{c}{CSG} \\
\cmidrule(lr){2-5} \cmidrule(lr){6-9} \cmidrule(lr){10-13}
& Rank1 & Rank5 & Rank10 & mAP & Rank1 & Rank5 & Rank10 & mAP & Rank1 & Rank5 & Rank10 & mAP \\
\midrule
CRRRO-BRO \cite{r7} & 17.8 & 34.6 & 48.1 & - & 23.3 & 54.0 & 69.8 & - & 10.4 & 25.8 & 37.5 & - \\
Covariance \cite{r6} & 38.0 & 61.0 & 73.1 & - & 26.5 & 52.5 & 66.0 & - & 16.5 & 34.1 & 47.9 & - \\
BSC+CM \cite{r8} & 58.6 & 80.6 & 87.4 & - & 32.0 & 59.1 & 72.3 & - & 24.6 & 38.5 & 55.1 & - \\
PREF \cite{r22} & 43.0 & 68.7 & 77.9 & - & 30.6 & 55.3 & 67.0 & - & 19.2 & 36.4 & 51.8 & - \\
LIMI \cite{r9} & 72.3 & 90.6 & 94.1 & - & 37.9 & 64.5 & 79.4 & - & - & - & - & - \\
DotGNN \cite{r11} & 74.1 & 90.1 & 92.6 & - & - & - & - & - & - & - & - & - \\
DotSCN \cite{r10} & 84.0 & \underline{95.1} & 96.3 & - & - & - & - & - & - & - & - & - \\
GCGNN \cite{r12} & 81.7 & 94.3 & 96.5 & - & 41.9 & 68.1 & 86.9 & - & - & - & - & - \\
SVIGR \cite{r23} & 87.8 & 92.7 & - & 89.2 & 46.2 & 71.8 & - & 42.1 & - & - & - & - \\
MGR \cite{r24} & 80.2 & 93.8 & 96.3 & - & 38.8 & 65.7 & 82.5 & - & 57.8 & 71.6 & 76.5 & - \\
MACG \cite{r25} & 84.5 & 95.0 & 96.9 & - & 45.1 & 70.4 & 84.9 & - & 63.2 & 75.4 & 79.7 & - \\
SOT \cite{r26} & 86.4 & 96.3 & \underline{98.8} & 91.3 & 58.8 & 70.6 & 88.2 & 59.9 & 91.7 & 96.5 & 97.6 & 90.7 \\
UMSOT \cite{r27} & \underline{88.9} & \underline{95.1} & \underline{98.8} & \underline{91.7} & \underline{64.7} & \underline{88.2} & \underline{99.9} & \underline{64.2} & \underline{93.6} & \underline{97.3} & \underline{98.3} & \underline{92.6} \\

\textbf{GCUM (Ours)} & \textbf{90.1} & \textbf{96.9} & \textbf{99.0} & \textbf{91.9} & \textbf{67.8} & \textbf{90.1} & \textbf{99.9} & \textbf{68.7} & \textbf{94.4} & \textbf{97.8} & \textbf{98.5} & \textbf{93.5} \\
\bottomrule
\end{tabular}
\label{table:combined}
\end{table*}

\section{Experiments}
\subsection{Datasets and Experimental Settings}
\textbf{Datasets.} We evaluate our GCUM method on three publicly available datasets: iLIDS-MCTS (IM) \cite{r7}, RoadGroup (RG) \cite{r9}, and CSG \cite{r25}. We use the cumulative matching characteristic (CMC) score and mean Average Precision (mAP) as our evaluation metrics.

\textbf{Implementation details.}
We randomly divide each dataset into a training set (70\%) and a test set (30\%). We employ a CLIP-ReID as the backbone network. The pedestrian images were resized to 256x128 as input. During training, the images were augmented with random horizontal flipping, cropping, and padding. We initiate the model training by gradually increasing the learning rate from $\text{5}\times \text{1}{{\text{0}}^{-\text{7}}}$ to $\text{5}\times \text{1}{{\text{0}}^{-\text{6}}}$ over the first 10 epochs. Then, the learning rate is reduced by a factor of 10 at the 30th and 50th epochs. Training was stopped after 80 epochs, with a batch size of 8. The optimizer used was Stochastic Gradient Descent (SGD), with a momentum of 0.8 and a weight decay of $\text{1e}^\textbf{-4}$.

\begin{table}[!t] 
\caption{Verifying the impact of different components on the overall performance of the model by gradually adding each module.}
\newcolumntype{C}{>{\centering\arraybackslash}X}
\begin{tabularx}{\linewidth}{lCCC}
\toprule
~ & \textbf{CSG} &\textbf{RG} &\textbf{IM} \\
\midrule
        Base &68.7 &75.4 &53.6 \\ 
        Base + GLA &76.9 & 81.5 & 61.2 \\ 
        Base + MVS &74.3 & 79.7 & 60.6 \\ 
        Base + GRCE &74.2 & 78.4& 58.4 \\ 
        Base + GLA + MVS &87.7 & 88.2 & 64.2 \\ 
        Base + GLA + MVS + GRCE &\textbf{94.4} &\textbf{90.1} &\textbf{67.8} \\
\bottomrule
\end{tabularx}
\label{table:3}
\end{table}

\subsection{Compared with Other Group Re-ID Methods}
Our method achieves the Rank-1 accuracies of 94.4\%, 90.1\% and 67.8\% on CSG, RoadGroup and iLIDS-MCTS datasets, respectively.

\textbf{Comparison with hand-crafted based methods.} 
Traditional manual methods rely on labor-intensive manual labeling, often resulting in low accuracy. In contrast, our approach outperforms most compared methods. On the largest CSG dataset, our method achieves a 75.2\% improvement in Rank-1 accuracy over PREF \cite{r22}. On the RoadGroup dataset, it improves Rank-1 accuracy by 31.5\% compared to BSC+CM \cite{r8}. Similarly, on the iLIDS-MCTS dataset, our method results in a 29.9\% increase in Rank-1 accuracy over PREF \cite{r22}.

\textbf{Comparison with deep learning based methods.}
Our method trained on a single dataset achieves state-of-the-art performance. Compared to the single-dataset trained UMSOT, our approach improves Rank-1 accuracy by 0.8\%, 1.2\%, and 3.1\% on the CSG, RoadGroup, and iLIDS-MCTS datasets, respectively. Even compared to the multi-dataset trained DotSCN, our method surpasses its accuracy without using additional datasets, highlighting our method's superiority.

\subsection{Ablation studies} 
We enhance the base model by adding the MVS, GLA, and Dual-Branch components. GLA connects individual person texts to form group texts, offering strong generalization to layout variations. MVS simulates changes in member numbers, providing broader group visual feature variations. The Dual-Branch component refines individual features for more comprehensive group features. These components effectively address variations in both members and layout. Integrating all three modules significantly improves performance, achieving Rank-1 accuracy of 94.4\%, 90.1\%, and 67.8\% on the CSG, RG, and IM datasets, respectively.

\begin{figure}[t]
\centering
\includegraphics[width=1\linewidth]{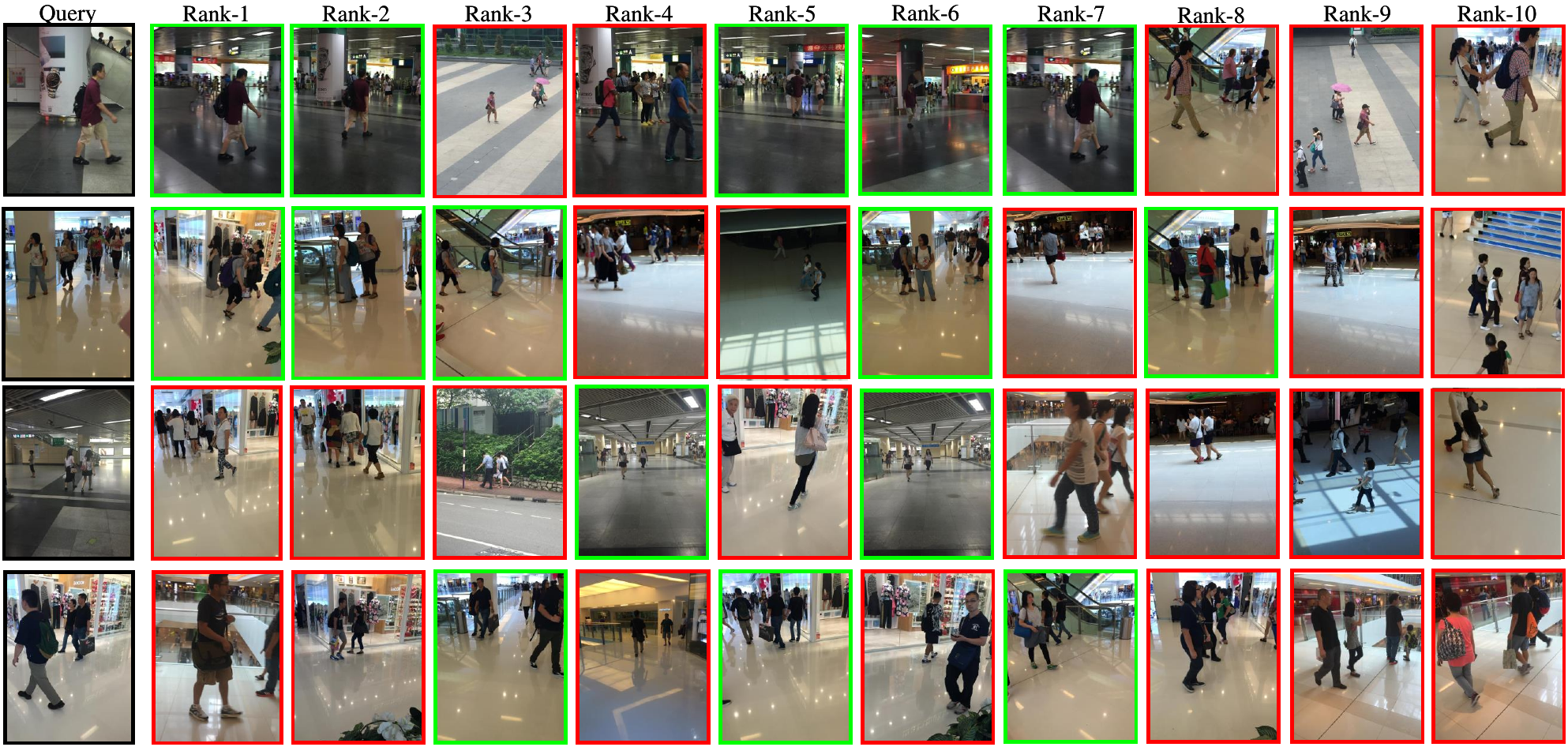}
\caption{The first image serves as the query, with the subsequent images representing the Rank-1 to Rank-10 retrieved results (from left to right). Green bounding boxes indicate correct matches, while red bounding boxes highlight incorrect matches.}
\label{Rerank}
\end{figure}

\subsection{Visualization}
Fig.\ref{Rerank} presents the retrieval results of several query samples. We observe that the proposed GCUM can accurately retrieve the same group images despite significant changes in member count and layout. This success is attributed to MVS, which simulates member variations, aiding GLA in generating text descriptions with strong adaptability to group changes. Furthermore, the GRCE captures more fine-grained group visual features and transfers text knowledge to visual features. Despite interference from different environments and resolution variations, our method consistently retrieves the correct group images.

\section{Conclusion}
In this paper, we are the first to introduce CLIP to Group ReID and propose a novel Group-CLIP Uncertainty Modeling (GCUM) method, which effectively overcomes the limitations of CLIP-ReID by utilizing uncertain group text descriptions. Secondly, we propose the Group Layout Adaptation (GLA) module and the Member Variant Simulation (MVS) module, which enhance the model's adaptability to changes in group structure. Additionally, the Group Relationship Construction Encoder (GRCE) is proposed to refine and aggregate mutli-modal features, ensuring robust performance across various scenarios. Experimental results on public datasets demonstrate that GCUM outperforms existing methods.

\bibliographystyle{IEEEtran}
\bibliography{ref}
\end{document}